\documentclass{article}
\usepackage{ijcai22}

\usepackage{times}

\usepackage{soul}
\usepackage{url}
\usepackage[colorlinks,citecolor=blue]{hyperref}
\usepackage[utf8]{inputenc}
\usepackage[small]{caption}
\usepackage{amsmath}
\usepackage{booktabs}
\usepackage{graphicx}
\usepackage{amsmath}
\usepackage{caption}

\usepackage{xcolor}
\definecolor{green}{RGB}{48,128,20}% colors
\usepackage{textcomp,booktabs}

\usepackage{graphics}
\usepackage{multirow}
\usepackage{pdfpages}
\usepackage{amsmath, amsfonts, amssymb, amsthm, xspace, color}
\usepackage{centernot}

\usepackage{graphicx} 
\usepackage{float} 
\usepackage{bm}

\usepackage{subfig}
\usepackage{hyperref}

\newcommand{\ie}{\emph{i.e.}\xspace} % that is
\newcommand{\eg}{\emph{e.g.\xspace}} % for example

\newcommand{\nop}[1]{}

\usepackage{latexsym}

\title{A Survey on Programmatic Weak Supervision}

\author{
Jieyu Zhang$^1$\thanks{These authors contributed equally to this work.}\and
Cheng-Yu Hsieh$^1$\footnotemark[1]\and
Yue Yu$^2$\footnotemark[1]\and
Chao Zhang$^2$\And
Alexander Ratner$^{1,3}$
\affiliations
$^1$University of Washington\\
$^2$Georgia Institute of Technology\\
$^3$Snorkel AI, Inc.
\emails
\{jieyuz2, cydhsieh, ajratner\}@cs.washington.edu\\
\{yueyu, chaozhang\}@gatech.edu
}

\begin{document}

\maketitle

\begin{abstract}
Labeling training data has become one of the major roadblocks to using machine learning. Among various weak supervision paradigms, programmatic weak supervision (PWS) has achieved remarkable success in easing the manual labeling bottleneck by programmatically synthesizing training labels from multiple potentially noisy supervision sources.  This paper presents a comprehensive survey of recent advances in PWS. In particular, we give a brief introduction of the PWS learning paradigm, and review representative approaches for each component within PWS's learning workflow. In addition, we discuss complementary learning paradigms for tackling limited labeled data scenarios and how these related approaches can be used in conjunction with PWS.
% We as well review recent datasets and applications of PWS in the literature. 
Finally, we identify several critical challenges that remain under-explored in the area to hopefully inspire future research directions in the field. 
% All the contents will be available at \url{https://github.com/JieyuZ2/Awesome-Weak-Supervision}. % may only for arxiv version

\end{abstract}

\section{Introduction}
% \red{deep learning, models are data hungry, data-centric ai (data is the problem), need to be creative about how to get huge volumes of training data, new weak supervision innovation}

During the last decade, deep learning and other representation learning approaches have achieved remarkable success, largely obviating the need for manual feature engineering and achieving new state-of-the-art scores across a broad range of data types, tasks, and domains.
However, they have largely done so via complex architectures that have required massive labeled training data sets.
% various fields and this unprecedented advance is largely attributed to the huge volumes of training data with manual annotations.
Unfortunately, manually collecting, curating, and labeling these training sets is often prohibitively time-consuming and labor-intensive.
% In other words, the strong models in current deep learning era are typically 
The data-hungry nature of these models has thus led to increased demand for innovative ways of collecting cheap yet substantial labeled training data sets, and in particular, labeling them.
% One of the greatest bottlenecks for deploying modern machine learning models in real-world applications is the need for substantial amounts of manually-labeled training data.

% \yue{add a para to include recent label-efficient methods} 
% 
To tackle the label scarcity bottleneck, a variety of classical approaches have seen a resurgence of interest.  
For instance, active learning (AL)~\cite{settles2009active,ren2021survey} aims to select the most informative samples to train the model with a limited labeling budget.
Semi-supervised learning~(SSL)~\cite{tarvainen2017mean,xie2020unsupervised} leverages a set of unlabeled data to improve the model's performance. 
% meta-learning~\cite{finn2017model,mishra2018simple} is proposed to learn from previous tasks, 
Transfer learning approaches~\cite{pan2009survey,wilson2020survey} pre-train a model or a set of representations on a source domain to enhance the performance on a different target domain.
% and zero-shot learning~\cite{wang2018zero,wang2019survey} aims to improve the model's generalizability over samples whose categories are unseen in training. 
However, these approaches still require a set of clean labeled data to achieve satisfactory performance, thus do not fully address the label scarcity bottleneck. 

To truly reduce the burdens of training data annotation, practitioners have resorted to cheaper sources of labels.
One classic approach is \emph{distant supervision} where external knowledge bases are leveraged to obtain noisy labels~\cite{hoffmann2011knowledge}.
There are also other options, including crowdsourced labels~\cite{yuen2011survey}, heuristic rules~\cite{Awasthi2020Learning}, feature annotation~\cite{mann2010generalized}, and others.
A natural question is: \emph{could we combine these approaches, and an even broader range of potential weak supervision inputs, in a principled and abstracted way?}

The recently-proposed programmatic weak supervision (PWS) frameworks provided affirmative answer to this question~\cite{Ratner16,ratner2017snorkel}. 
Specifically, in PWS, users encode \emph{weak supervision sources}, \eg, heuristics, knowledge bases, and pre-trained models, in the form of \emph{labeling functions (LFs)}, which are user-defined programs that each provide labels for some subset of the data, collectively generating a large set of training labels.

The labeling functions are usually noisy with varying error rates and may generate conflicting labels on certain data points. 
To address these issues, researchers have developed \emph{label models}~\cite{Ratner16,Ratner19,fu2020fast,Varma2019multi} which aggregate the noisy votes of labeling functions to produce training labels.
Then, the training labels is in turn used to train an \emph{end model} for downstream tasks.
These two-stage methods mainly focus on the efficiency and effectiveness of label model, while maintaining the maximal flexibility of the end model.
% One exception is COSINE~\cite{yu-etal-2021-fine}, which aims to design a better end model by bootstraping over data uncovered by labeling functions in a self-training manner.
% Another work~\cite{li2021bertifying} trains the label model and the end model alternately using each other's output.
% The motivation is that due to the low coverage of labeling functions, there may exist an ignorable portion of data that are not covered by labeling functions and are usually dropped in practice, COSINE in contrast utilizes these uncovered data in a self-training manner.
In addition to the two-stage methods, later researchers also explored the possibility of coupling the label model and the end model in an end-to-end manner~\cite{ren2020denoising,lan2020connet}. 
We refer to these one-stage methods as \emph{joint models}.
An overview of weak supervision pipeline can be found in Fig.\ref{fig:overview}.

In addition, these LFs often have clear dependencies among them~\cite{Ratner16} and therefore it is crucial to specify and take into consideration the appropriate dependency structure~\cite{MisspecificationInDP}. However, manually specifying the dependency structure would bring extra burden to practitioners; to reduce human efforts, researches have attempted to learn the dependency structure automatically~\cite{Bach2017LearningTS,Varma2017InferringGM,Varma2019LearningDS}. 
Very recently, researchers have also explored the possibility of generating these LFs automatically~\cite{varma2018snuba} or interactively~\cite{boecking2021interactive}.

In this paper, we present the first survey on PWS to introduce its recent advances, with special focus on its formulations, methodology, applications, and future research directions. 
We organize this survey as follows: after a brief introduction of PWS in Sec.~\ref{sec:background}, we review approaches for each component within a standard PWS workflow, namely, the label model (Sec.~\ref{sec:label_model}), end model (Sec.~\ref{sec:end_model}), and joint model (Sec.~\ref{sec:joint_model}).
Then, we briefly address complementary approaches for the limited label scenario and how they interact with PWS.
Finally, we discuss the % datasets and application (Sec.~\ref{sec:dataset}) and
challenges and future directions (Sec.~\ref{sec:future}).
We hope that this survey can provide a comprehensive review for interested researchers, and inspire more research in this and related areas.

\section{Preliminary}
\label{sec:background}

\begin{table*}[t]
    \centering
    \caption{Comparisons among existing methods for each component of the PWS pipeline. *: NPLM and PLRM are able to utilize new types of LFs as described in Sec~\ref{sec:label_model}.}
    \scalebox{0.58}{
    \setlength{\tabcolsep}{2em}
    \begin{tabular}{ l l l  c c c c c}
        \toprule
        \multirow{2}{*}{\textbf{Module}} & \multirow{2}{*}{\textbf{Target Task}} &  \multirow{2}{*}{\textbf{Method}} &
        \multicolumn{4}{c}{\textbf{Input}}  \\ \cmidrule(lr){4-7}
         
        &  &  & \textbf{X}  &  \textbf{P(Y)} & \textbf{Additional Information} & \textbf{LF dependency}\\
         \toprule
         
    \multirow{12}{*}{Label Model} & \multirow{6}{*}{Classification} &  Data Programming~\cite{Ratner16} &  &  & & \checkmark\\
    & & MeTaL~\cite{Ratner19} & & \checkmark & & \checkmark\\
    & & FlyingSquid~\cite{fu2020fast} &  & \checkmark & & \checkmark\\
    
    % & PGMV~\cite{mazzetto:aistats21} &  & & Error Rate of LFs &  \\
    & & CAGE~\cite{chatterjee2020robust} &  & & User-provided Quality of LFs & \\
    & & NPLM$^{*}$ ~\cite{yu2021learning} &  & &  \\
    & & PLRM$^{*}$ ~\cite{zhang2021creating} &  & & \\\cmidrule(lr){2-7}

    & \multirow{4}{*}{Sequence Tagging} &   Dugong~\cite{Varma2019multi} &  & \checkmark & & \checkmark\\
    & & HMM~\cite{lison2020named} & & \checkmark \\
    & & Linked HMM~\cite{safranchik2020weakly} &  & & Linking Functions & \\
    & & CHMM~\cite{li2021bertifying} & \checkmark \\   \cmidrule(lr){2-7}
     
    & Classification, Ranking, Regression & \multirow{2}{*}{UWS~\cite{shin2021universalizing}} &  &  & & \multirow{2}{*}{\checkmark}\\
    & Learning in Hyperbolic Manifolds\\
    
    \midrule
    \multirow{1}{*}{End Model} & \multirow{1}{*}{Classification} & COSINE~\cite{yu-etal-2021-fine} & \checkmark \\
    
    \midrule
    \multirow{9}{*}{Joint Model} & \multirow{7}{*}{Classification}  & {Denoise~\cite{ren2020denoising}} & \checkmark  \\
    & & WeaSEL~\cite{cachay2021endtoend} & \checkmark & \checkmark \\
    & & ALL~\cite{arachie2019adversarial} & \checkmark & & Error Rate of LFs &  \\
    & & AMCL~\cite{mazzetto:icml21} & \checkmark  & & Set of Labeled data& \\
    
    & & ImplyLoss~\cite{Awasthi2020Learning} & \checkmark & & Exemplar Data of LFs & \\
    & & ASTRA~\cite{karamanolakis2021self} & \checkmark & & Set of Labeled data& \\ 
    & & SPEAR~\cite{maheshwari2021semi} & \checkmark & & Set of Labeled data &\\\cmidrule(lr){2-7}
    
     &  \multirow{2}{*}{Sequence Tagging}  & ConNet~\cite{lan2020connet} & \checkmark  \\
     & & DWS~\cite{parker2021named} & \checkmark \\
    \bottomrule
    \end{tabular}
    }
    \label{tab:methods}
\end{table*}

% We first give some background on programmatic weak supervision (PWS) at a high level.
% In PWS, users create multiple weak supervision sources that assign noisy labels to data.
% Critically, these weak supervision sources can vote or abstain on individual data points; this lets users express high-precision signals without requiring them to have high recall as well.
% The goal of PWS is to produce an end machine learning model based on the source votes for downstream task.
% Generally, existing weak supervision methods can be put into two categories: two-stage methods and one-stage methods.

Now, we formally define the setting of PWS.
We are given a dataset $D$ with $n$ data points and the $i$-th data point is denoted by $X_i \in \mathcal{X}$.
For each $X_i$, there is an unobserved true label denoted by $Y_i \in \mathcal{Y}$.
Let $m$ be the number of sources $\{S_j\}_{j\in[m]}$, each assigning a label $\lambda_j \in \mathcal{Y}$ to some $X_i$ to vote on its respective $Y_i$,
or abstaining ($\lambda_j = -1$).
In addition, some methods could handle the dependencies among sources by inputting the dependency graph of sources $G_{dep}$.
For concreteness, we follow the general convention of PWS~\cite{Ratner16} and refer to these sources as \emph{labeling functions (LFs)} throughout the paper.
The goal is to apply $m$ LFs to the unlabeled dataset $\bm{X}=[X_1, X_2, \ldots, X_n]$ to create an $n \times m$ label matrix $L$,
and to then use $L$ and $\bm{X}$ to produce an end machine learning model $f_w : \mathcal{X} \rightarrow \mathcal{Y}$.

\begin{figure}[t]
  \centering
  \includegraphics[width=0.49\textwidth]{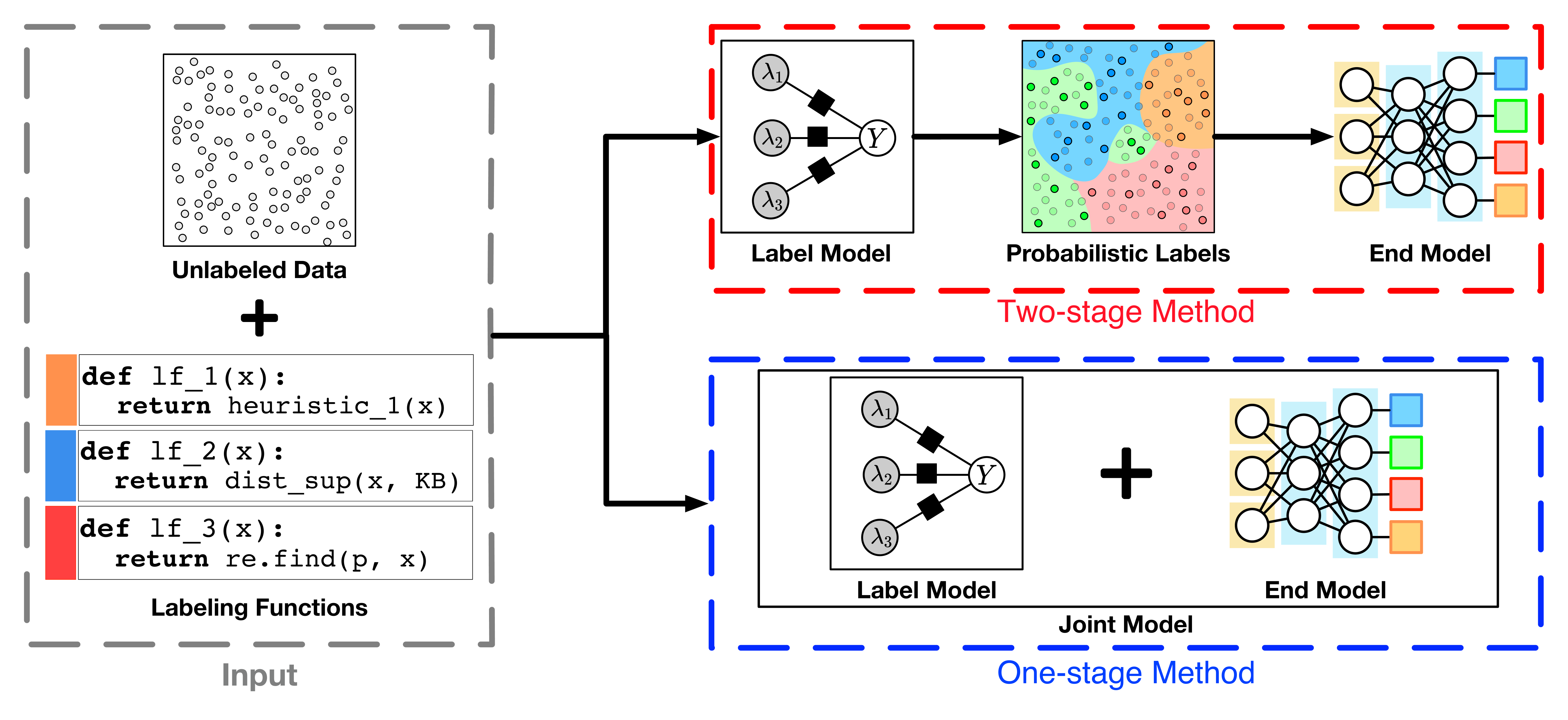} 
  \caption{An overview of PWS pipeline~\protect\cite{zhang2021wrench}.}
  \label{fig:overview}
\end{figure}

In general, PWS methods could be classified into two categories as shown in Fig~\ref{fig:overview}:

\paragraph{Two-stage Method.}

A two-stage method works as follows.
In the first stage, a \textit{label model} is used to combine the label matrix $L$ into either probabilistic soft training labels or one-hot hard training labels, which are in turn used to train the desired \emph{end model} in the second stage.
We review the label models and end models in the literature separately.

% \paragraph{Label Model}

% Most of the studies of two-stage methods focus on developing new label models, leaving the end model a flexible choice based on downstream tasks. 
% The label model can be as simple as Majority Voting, which is parameter-free.
% More advanced methods often treat the ground truth label $Y$ as latent variable and model the joint distribution of $Y$ and $\{\lambda_j\}_{j\in[m]}$ in probabilistic graphical models (PGMs).
% Then, the learned PGM is used to infer training labels for the unlabeled dataset.
% We elaborate these methods below.

\paragraph{One-stage Method.}
The one-stage methods attempt to train a label model and end model simultaneously. Specifically, they usually design a neural network for label aggregation while utilizing another neural network for final prediction.
These approaches offer a more straightforward way for tackling weak labels.
% as two modules are jointly trained in an end-to-end manner.
% Moreover, different from two-stage methods that generate training labels in a context-free manner, one-stage methods encode contextual information of training samples via neural networks. 
We refer to the model designed for one-stage method as a \emph{joint model}.

\section{Labeling Functions}
At the core of PWS are the labeling functions (LFs) that provide potentially noisy weak labels that fuel the entire learning pipeline. In this section, we provide an overview over the popular types of LFs, how they are generally developed, and the potential dependency structure among the LFs.

\subsection{Labeling Function Types}
In PWS, users encode different weak supervision sources into LFs, each of which noisily annotates a subset of data points. While an LF can be as general as any function $\lambda: \mathcal{X} \to \mathcal{Y} \cup \{-1\}$ that takes as input a data point and either outputs a corresponding label or abstain, we introduce the most common types of LFs used in practice.

\subsubsection{User-written Heuristics}
In practical applications, users generally have domain knowledge about the target learning task of interest. One common type of LF is to express the domain knowledge into heuristic labeling rules that associate corresponding labels to the data points. For example, in text applications, users write keyword- or regex-based LFs that assign corresponding labels to the data points that contains the keyword or matches the specified regular expression~\cite{ratner2017snorkel,meng2018weakly,Awasthi2020Learning}. In image applications, users write LFs that provide labels to the image inputs containing specific objects, or possessing some user-specified visual/spatial properties~\cite{Varma2017InferringGM,chen2019scene,fu2020fast}.

\subsubsection{Existing Knowledge}

\paragraph{Knowledge Bases.} Oftentimes, external knowledge bases can be used to provide weak supervision over the learning task of interest, commonly known as the distant supervision approach~\cite{hoffmann2011knowledge,liang2020bond}.
For example, in a relation extraction task to identify mentions of spouse
relationships in news article, \cite{ratner2017snorkel} writes LFs that match the text inputs against the knowledge base DBPedia\footnote{\url{https://www.dbpedia.org/}} to search for known spouse relationships.

\paragraph{Pre-trained Models.} Existing pre-trained models from a related task can be used as LFs to provide weak labels. For example, in a product classification task at Google, \cite{bach2019snorkel} leverages existing semantic topic model to identify contents irrelevant to the category of products of interest. In \cite{zhang2021creating}, pre-trained image classification model that has a different output label space from the target classification task is used as LFs to provide indirect weak supervision for the learning task of interest.

\paragraph{Third-party Tools.}
To collect weak labels cheaply, there are several existing third-party tools available that can serve as LFs. 
For example, for review sentiment analysis, users can simply use TextBlob\footnote{\url{https://textblob.readthedocs.io/en/dev/}} to assign labels for each  review. Take named entity recognition (NER) as another example, there are several tagging tools such as spaCy\footnote{\url{https://spacy.io/}}, NLTK\footnote{\url{https://www.nltk.org/}}, etc, and \cite{lison2020named} adopt them as LFs for weakly-supervised NER task. 
Note that, the above tools are not perfect, as the weak labels generated via their outputs contains much noise.
% third party tools
% \footnote{https://textblob.readthedocs.io/en/dev/}

\subsubsection{Crowd-sourced Labels.} Crowd-sourcing is the classic and well-studied approach of obtaining less accurate label annotations from non-expert contributors with lower annotation cost~\cite{DawidSkene,yuen2011survey}. In the PWS setting, each crowd-sourcing contributor can be represented as an LF that noisily annotates the data points~\cite{ratner2017snorkel,lan2020connet}. For example, in a weather sentiment classification task, each crowd-source contributor---who grades the sentiment of tweets relating to the weather into five different categories---is considered as a LF.

\subsection{Labeling Function Generation}
In the PWS learning paradigm, the first and foremost step is to create a set of LFs that are used to generate the weak labels for learning the subsequent models. In practice, the LFs are typically developed by subject matter experts (SMEs)
who have adequate knowledge about the task of interest. When developing LFs, in addition to leveraging existing domain knowledge, SMEs usually refer to a small subset of data points sampled from the unlabeled set, called the \textit{development set}, to extract further task/dataset-specific labeling heuristics that complement the pre-existing domain knowledge~\cite{ratner2017snorkel}.
This process of LF development could sometimes be challenging and time-consuming even for domain experts. For example, it often requires SMEs to explore a considerable amount of development data to generate ideas for LFs~\cite{varma2018snuba,darwin,boecking2021interactive}.
As a result, researchers have recently aim to reduce the amount of efforts spent in designing weak supervision sources through three main directions, namely, \textit{automatic generation}, \textit{interactive generation}, and \textit{guided generation} of LFs.

\paragraph{Automatic Generation.}
One direction to alleviate the burden in designing LFs in PWS paradigm is to automate the process of LF development. \cite{varma2018snuba} propose a system, Snuba, that generates LFs automatically by learning weak classification models on a small set of labeled dataset.
TALLOR~\cite{TALLOR} takes as input an initial set of seed LFs that are generally simpler, and automatically learn more accurate compound LFs from multiple simple labeling rules.
Similarly, GLaRA~\cite{glara} learns to augment a set of seed LFs automatically by exploiting the semantic relationships between candidate and seed LFs through a graph-based model.
Notably, while we refer to this line of methods as ``automatic generation'' approaches, they do require a minimum amount of initial supervision, either in the form of small labeled set or seed LFs.

% Notably, \cite{zhang2021wrench} leverages all the golden training labels to automatically generate LFs but only for study purpose.

% nature language \cite{hancock-etal-2018-training}

\paragraph{Interactive Generation.}
In contrast to fully automating the generation of LFs \textit{after} given a seed supervision set, interactive generation approaches cast LF development as an interactive process where users are iteratively queried for feedback used in discovering useful LFs from a large set of candidates~\cite{darwin,boecking2021interactive}.
Specifically, in Darwin~\cite{darwin} and IWS~\cite{boecking2021interactive}, a set of candidate LFs is first generated based on $n$-grams or context-free grammar information. Then, in each iteration, the user is queried to annotate whether a presented LF, proposed by the system, is useful or not (i.e., better than random accuracy). Based on the feedback provided in each iteration, the systems learn to adapt and identify a set of high-precision LFs from the candidate set, which is used as the final set of LFs in the PWS learning pipeline.
Compared to the standard active learning approaches which relies on instance-level annotations, the interactive generation approaches are shown to achieve better performance with lower annotation costs.

\paragraph{Guided Generation.}
Based on the current workflow of LF development where SMEs write LFs by looking at a small development set of data, guided-generation approaches aim to assist the users in developing LFs by intelligently curate the development set in order to efficiently \textit{guide} SMEs in exploring the data and developing informative LFs that could lead to strong resultant models~\cite{cohen-wang2019interactive}. 
The idea resembles traditional active learning~\cite{settles2009active} in the sense that the goal is to strategically select data points from the unlabeled set and solicit informative supervision from the users, except that the supervision is provided at the functional-level (i.e., LFs) instead of individual label level.

\section{Label Model}
\label{sec:label_model}
The multiple LFs we have for a given dataset often overlap and conflict with each other. In PWS, \emph{label model} is used to integrate the LFs' output predictions into probabilistic labels, aiming to accurately recover the unobserved ground truth labels.
Till now, various label models have been proposed and most of them are based on probabilistic graphical models.
It is worthwhile to note that LFs developed in practice often exhibit statistical dependency among each other~\cite{Ratner16,MisspecificationInDP}.
% Users commonly introduce correlated LFs by engineering new ones that are used either to \textit{reinforce} or to \textit{fix} existing LFs, leading to various types of dependencies and structures among the weak supervision sources~\cite{Ratner16,MisspecificationInDP}.
Incorporating the dependency information into the label model has been shown to be critical to the model's ability to correctly estimate the latent ground truths~\cite{Ratner16,Bach2017LearningTS,Varma2017InferringGM,MisspecificationInDP}. However, not all label models take into account the LF dependency structure when aggregating the LFs' votes, where some approaches simply assume conditional independence between the LFs.

In this section, we first discuss general approaches used to incorporate LF dependency in label model. Then, we introduce more in detail different existing label models, categorized by their target learning tasks, with discussion on how the LF dependency is handled in some of the approaches.

% sometimes taking the dependency structure of the LFs into consideration as illustrated above. Till now, various label models have been proposed and most of them are based on probabilistic graphical models.

\subsection{LF Dependency Structure}
Earlier work on PWS rely on users to manually specify the dependency structure among the LFs~\cite{Ratner16}. For example, users could specify two LFs to be \textit{similar}; or one LF to be \textit{fixing} or \textit{reinforcing} another; or two LFs are \textit{exclusive}.
Nevertheless, as manually specifying such dependency structure is generally hard for users, researchers have recently turned to \textit{learning} or \textit{inferring} the dependency structure automatically without user supervision.
To automatically \textit{learn} the dependency structure, \cite{Bach2017LearningTS} proposes to maximize the $\ell_1$-regularized marginal pseudo-likelihood of a factor graph with high-order dependencies and select the dependencies that have non-zero weights; \cite{Varma2019LearningDS} exploits the sparsity of label model and leverages robust PCA technique to capture the underlying dependency structure.
On the other hand, instead of \textit{learning} the structure from the observed labels, \cite{Varma2017InferringGM} proposes an alternative approach that \textit{infers} the relations between different LFs by statically analyzing the source code of the LFs.

Having the dependency structure on hand, whether manually specified or automatically learned/inferred, a prevailing approach to incorporate the dependency information into the label model is to embed the dependency relationships into label models, which are typically graphical models, through factor functions~\cite{Ratner16,shin2021universalizing} or graph structure~\cite{Ratner19,fu2020fast,Varma2019multi}. In the following subsections, we introduce the label models for different learning tasks in more detail, and provide an overview of these methods in Table~\ref{tab:methods}.

% \paragraph{DP~\protect\cite{Ratner16}:} 
\subsection{Label Model for Classification} For classification problems, majority voting (MV) is the most  straight-forward approach for aggregating different LFs, as it simply uses the consensus from the multiple LFs to obtain more reliable labels without introducing any trainable parameters.
% Dawid-Skene (DS) model estimates the accuracy of each LF with expectation maximization (EM) algorithm by assuming a naive Bayes distribution over the LFs’ votes and the latent ground truth.
Crowdsourcing models~\cite{DawidSkene,dalvi2013aggregating,raykar2010learning,khetan2018learning} usually leverage the expectation maximization (EM) algorithm to estimate the accuracy of each worker as well as infer the latent ground truth labels, which can also be applied here when we regard each LF as a worker. 
Apart from these approaches, we review several label models tailored for PWS problems. 
These label models are all based on probabilistic graphical model and aim to maximize the probability of observing the outputs of LFs.
Specifically, they share an optimization problem as following:

\begin{equation}
\label{eq:ci}
\max_{\theta} P(L; \theta) = \sum_{Y}P(L, Y; \theta) ~.
\end{equation}

The key differences among existing label models are the way they parameterize the joint distribution $P_{\theta}(L, Y)$ and how the parameters are estimated.
In particular, Data programming (DP)~\protect\cite{Ratner16} models the distribution  $P(L, Y; \theta)$ as a factor graph. It is able to describe the distribution in terms of  pre-defined factor functions, which reflects the dependency of any subset of random variables and are also used to encode the dependency structure of LFs. 
The log-likelihood is optimized by SGD where the gradient is estimated by  Gibbs sampling.
MeTaL~\protect\cite{Ratner19}, instead, models the distribution via a Markov Network and recover the parameters via a matrix completion-style approach.
% Notably, it requires label prior as input.
% \paragraph{FlyingSquid~\protect\cite{fu2020fast}:} 
Later on, FlyingSquid~\cite{fu2020fast} is proposed to accelerate the learning process for binary classification problems. It models the distribution as a binary Ising model, where each LF is represented by two random variables, and a Triplet Method is used to recover the parameters and therefore no learning is needed.
Notably, the latter two methods encode the dependency structure of LFs into the structure of the graphical model and require label prior as input.

Additionally, researchers have attempted to extend the scope of usable LFs.
% by developing label models that could leverage other types of LFs. 
CAGE~\cite{chatterjee2020robust} extends the existing label models to support continuous LFs. In addition, it leverages user-provided quality for LFs  
% quality guides
% It extends the generative model through quality guides,
to increase the training stability and making it less sensitive to initialization.
Moreover, NPLM~\cite{yu2021learning} enables users to utilize partial LFs that output a subset of possible class labels and PLRM~\cite{zhang2021creating} allows the usage of indirect LFs that only predict unseen but related class;
both works are built on probabilistic graphical model similar to~\cite{Ratner16} and greatly expand the scope of usable LFs in PWS.
% which makes it much faster than data programming and MeTal.
% Notably, FlyingSquid is designed for binary classification and the author suggested applying a one-versus-all reduction repeatedly to apply the core algorithm.

% \paragraph{Modeling LF Dependency.}
% Notably, the DP model~\protect\cite{Ratner16} is able to handle arbitrary LF dependency structure by explicitly encoding the dependency as factor functions in a factor graph. Similarly, \cite{Ratner19,fu2020fast} consider LF dependencies by embedding the dependency structure in Markov network-based label models.

\subsection{Label Model for Sequence Tagging} Sequence tagging problems are more complex since there are dependencies among consecutive tokens. To model such properties, Hidden Markov Models (HMM)~\cite{baum1966statistical} have been proposed, which represent true labels as latent variables and inferring them from the independently observed noisy labels through expectation-maximization algorithm \cite{welch2003hidden}. \cite{lison2020named} directly apply HMM for named entity recognition task and \cite{safranchik2020weakly} propose Linked-HMM to incorporate unique linking rules as an adjunct supervision source additional to general weak labels on tokens. Moreover, Conditional hidden Markov model (CHMM)~\cite{li2021bertifying} substitutes the constant transition and emission matrices by token-wise counterpart predicted from the BERT embeddings to model  the evolve of true labels in a context-dependent manner.
% \paragraph{PGMV~\protect\cite{mazzetto:aistats21}:} 
% This method only works for binary classification and finds a subset of weak supervision sources whose majority vote achieves high accuracy with respect to the worst-case distribution of the output of the weak supervision
% sources. Notably, this worst-case distribution is constrained by using statistics computed on the weak supervision sources (individual error rates and pairwise differences).
Another characteristics for sequence tagging problems is that the supervision can be provided at different resolutions (e.g. frame, window, and scene-level for videos). To integrate them together, Dugong~\cite{Varma2019multi} has been propose to assign probabilistic labels for data with graphical models. Dugong also accelerates the inference speed with SGD based optimization techniques.
Finally, as shown in~\cite{zhang2021wrench}, label models for classification task could also be applied on sequence tagging problem with certain adaptations.

% \paragraph{Modeling LF Dependency.}
% For modeling the LF dependency in sequence tagging task, \cite{Varma2019multi} encode the dependency structure into the label model similar to the technique used in \cite{Ratner19} for classification task.

\subsection{Label Model for General Learning Tasks} Very recently, UWS~\cite{shin2021universalizing} goes beyond the traditional tasks and generalizes PWS frameworks to handle more kinds of tasks including ranking, regression, and learning in hyperbolic manifolds with an efficient method-of-moments approach in the embedding space.
% \paragraph{Dugong~\protect\cite{Varma2019multi}:} Dugong is a framework that  integrates weak supervision sources at different resolutions (e.g. frame, window, and scene-level for videos) to assign probabilistic
% labels for data with graphical models. 
% They also accelerate the inference speed with SGD based optimization techniques.

% \paragraph{HMM~\protect\cite{lison2020named}:}
% Hidden Markov models %(HMMs, shown in Figure~\ref{fig:hmm} circled part)
%  represent true labels as latent variables and inferring them from the independently observed noisy labels through expectation-maximization algorithm \cite{welch2003hidden}.

% \paragraph{Linked-HMM~\protect\cite{safranchik2020weakly}:}
% Linked hidden Markov models (linked HMMs) incorporates unique linking rules as an adjunct supervision source additional to general token labels.
% The linking rules model the probabilities of two adjacent tokens belonging to the same named entity span.

% \paragraph{CHMM~\protect\cite{li2021bertifying}:}
% Conditional hidden Markov model (CHMM)  substitutes the constant transition and emission matrices by token-wise counterpart predicted from the BERT embeddings.
% The token-wise probabilities are representative in modeling how the true labels evolve according to the input tokens.

\section{End Model}
\label{sec:end_model}
After obtaining the probabilistic labels, the end model is used to train a discriminative model on downstream tasks. 
Since the probabilistic training labels derived from the label model may still contain noise, \cite{ratner2017snorkel} suggests using a noise-aware loss as the training objective for the end model. 
However, one drawback for such end models are that they are usually trained only on the data covered by weak supervision, but there may exist an ignorable portion of data that are not covered by any LFs.
% \paragraph{COSINE~\protect\cite{yu-etal-2021-fine}:}
Motivated by this, COSINE~\cite{yu-etal-2021-fine} designs a better end model by leveraging the data uncovered by LFs. Specifically, it utilizes these uncovered data in a self-training manner and generates pseudo labels for each unlabel data.
% The motivation is that due to the low coverage of LFs, there may exist an ignorable portion of data that are not covered by LFs and are usually dropped in practice, COSINE in contrast utilizes these uncovered data in a self-training manner.
Apart from the above methods, other approaches designed for learning with noisy labels~\cite{song2020learning} can also be utilized as end models.

\section{Joint Model}
\label{sec:joint_model}
The traditional pipeline for PWS usually trains the label model and end model separately, in contrast, \emph{joint model} aims to train the label model and the end model in an end-to-end manner, allowing them to enhance each other mutually.
In addition, the joint model usually leverages neural network as label model instead of aforementioned statistical label model; such a design choice not only facilitates the co-training of label model and end model, but also reflects the motivation of considering data feature during the training of label model, leading to a instance-dependent label model, \ie, $P(L, Y|X; \theta)$. 
% simultaneously. The two components are often integrated into a co-training framework, allowing them to enhance each other mutually.
% \red{need more motivations here, use the term neural label model}

% \paragraph{Modeling LF Dependency.}
As opposed to statistical label models (Sec.~\ref{sec:label_model}) that \textit{explicitly} incorporate LF dependency through the graph structure of underlying graphical models, it is observed that neural network based joint models are able to \textit{implicitly} capture the dependencies among the LFs in the learning process~\cite{cachay2021endtoend}. However, existing joint models generally cannot incorporate pre-given dependency structure.

\subsection{Joint Model for Classification}
% y maximizing its agreement with probabilistic labels generated by reparameterizing prior probabilistic posteriors with a neural network
Denoise~\cite{ren2020denoising} and WeaSEL~\cite{cachay2021endtoend} first reparameterize prior probabilistic posteriors with a neural network, then assign scores for each PWS source for aggregation. After that, the posterior network and the end model are trained simultaneously to maximize the agreement between them. 
\cite{arachie2019adversarial,mazzetto:icml21} both formulate the weakly supervised classification problems as a constrained min-max optimization problem, and ALL~\cite{arachie2019adversarial} learns a prediction model that has
the highest expected accuracy with respect to an adversarial labeling of an unlabeled dataset, where this labeling must satisfy error constraints on the weak supervision sources. 
Differently, AMCL~\cite{mazzetto:aistats21} constructs the constraints based on the expected loss within a small set of clean data.
% \paragraph{Denoise~\protect\cite{ren2020denoising}:} Denoise adopts an attention network to aggregate over weak labels, and use a neural classifier to leverage the data feature. These two components are jointly trained in an end-to-end manner.
% \paragraph{WeaSEL~\protect\cite{cachay2021endtoend}:} WeaSEL is an end-to-end  WS framework. Specifically, they use a neural encoder network to produce accuracy scores on each WS source for aggregation, then it trains the encoder and downstream model simultaneously to maximize the agreement between them.
% \paragraph{ALL~\protect\cite{arachie2019adversarial}:} Adversarial Label Learning (ALL) learns a prediction model that has the highest expected accuracy with respect to an adversarial labeling of an unlabeled dataset, where this labeling must satisfy error constraints on the weak supervision sources. In particular, it formulates a constrained min-max optimization problem and solves it via  augmented Lagrangian method.
% \paragraph{AMCL~\protect\cite{mazzetto:icml21}:} Similar to ALL, AMCL aims to solve a constrained min-max optimization problem in order to learn a model that perform the best with regard to an adversarial labeling. Differently, AMCL construct the constraints based on the expected loss with a small set of clean data. In addition, AMCL also provides theoretical performance guarantees for the learned model.

% \paragraph{Semi-PWS for Classification}
To denoise LFs more effectively, several methods propose to use a small number of labeled data in training. 
% We named such methods as Semi-PWS approaches since their setups are also close to semi-supervised learning. 
% \paragraph{ImplyLoss~\protect\cite{Awasthi2020Learning}:} 
ImplyLoss~\cite{Awasthi2020Learning} jointly train a rule denoising network based on exemplars for each label, as well as a classification model with a soft implication loss.
SPEAR~\cite{maheshwari2021semi} extends ImplyLoss by designing additional loss functions  on both labeled and unlabeled data and encourages the consistency between the two models. 
In addition, ASTRA~\cite{karamanolakis2021self} adopts self-training for PWS with a teacher-student framework. 
% \paragraph{ASTRA~\protect\cite{karamanolakis2021self}:} 
%ASTRA is a semi-weakly supervised method based on a teacher-student framework.
The student model is initialized with a small number of labeled data and generates pseudo-labels for instances not covered by LFs, while 
the teacher model combines LFs with the output from the student model for the final prediction. 
% The self-trained student model predictions 
% \paragraph{CAGE~\cite{chatterjee2020robust}}
% \paragraph{SPEAR~\protect\cite{maheshwari2021semi}:} SPEAR jointly trains a classification model based on raw features and a label aggregation model based on LFs in a semi-supervised manner.  Moreover, it designs several loss functions including cross-entropy on both labeled and unlabeled data as well as the consistency loss to encourage the similar output between the feature model and the label aggregation model. 

\subsection{Joint Model for Sequence Tagging}
For sequence tagging problem, Consensus Network (ConNet)~\cite{lan2020connet}
% \paragraph{ConNet~\protect\cite{lan2020connet}:}
% Consensus Network (ConNet)  adopts a two-stage approach for learning with multiple  supervision signals. In the decoupling phase, it 
trains BiLSTM-CRF \cite{ma2016end} with multiple CRF layers for each labeling source individually. Then, it aggregates the CRF transitions with attention scores conditioned on the quality of LFs and outputs a unified label sequence.
% \paragraph{DWS~\protect\cite{parker2021named}:}
DWS~\cite{parker2021named} uses a CRF layer to capture statistical dependencies among tokens, weak labels and latent true labels. Moreover, it adopts hard EM algorithm for model training: in the E-step, it finds the most probable labels for the given sequence; in the M-step, it maximizes the probability for the corresponding labels.

\section{Complementary Approaches}

In this section, we briefly describe how PWS can be connected to or combined with complementary machine learning approaches that also aim to deal with the label scarcity issue.

\paragraph{Active Learning.}
Active learning (AL) attempts to handle the label scarcity issue by interactively annotating the most informative samples to achieve good performance.
As complementary approach, PWS could be utilized to improve AL. For example, \cite{Mallinar2020IterativeDP} expands the initial labeled set in AL by querying labels for those which are the most relevant to existing labeled ones based on LFs, and \cite{Nashaat2018HybridizationOA} applied PWS to generate initial noisy training labels to improve the efficiency of a later active learning process.
On the other hand, AL could in turn help PWS: \cite{biegel2021active} asks experts to provide labels for which the label model is most likely to be mistaken, and Asterisk \cite{nashaat2020asterisk} employs AL to enhance the label model and proposed a selection policy based on the estimate accuracy of LFs and the output of label model.

\paragraph{Transfer Learning.}
Transfer learning (TL), which adapts a trained model to the new tasks and consequently tends to require less labeled data than training from scratch, has recently attract increasing attention, especially for the great success of fine-tuning huge pretrained models with few labels. We note that TL and PWS are orthogonal to each other and could be combined together to achieve the best performance, since TL could reduce but not eliminate the demand of labeled data, which could be offered by PWS.
Indeed, current state-of-the-art PWS methods usually rely on fine-tuning pretrained models with labels produced by label model~\cite{zhang2021wrench}. 

\paragraph{Semi-Supervised Learning.}
Semi-supervised Learning (SSL) aims to train the model with a small amount of labeled data with a large amount of unlabeled data. 
The idea of leveraging unlabeled data to improve training has also been applied to PWS methods as \cite{karamanolakis2021self,yu-etal-2021-fine} use self-training to bootstrap over unlabeled data.
% \red{mention cosine and other related work}
Moreover, \cite{xu2021dp} improves SSL by leveraging the idea of PWS; specifically, they use the labeled data to generate LFs that are in turn used to  annotate the unlabeled data, finally the model is trained on the whole dataset with provided or synthesized labels.
To sum up, SSL and PWS are also complementary and future works include developing more advanced methods to combine clean labels and weak labels together to further boost the performance.

\section{Challenges and Future Directions}
\label{sec:future}

\paragraph{Extend to More Complex Tasks.} The majority of the PWS methods only support classification or sequence tagging tasks, while there are a variety of tasks that require high-level reasoning over concepts such as question answering~\cite{rajpurkar2016squad}, navigation~\cite{gupta2017cognitive} and scene graph generation~\cite{ye2021linguistic}, and curating labeled data for these tasks requires even more  human efforts.
Moreover, in these tasks, the input data may come from multiple modalities including  text, image and tables, while the current PWS methods only consider LFs with one specific modality. 
% However, for many other tasks such as navigation and visual question answering, it is essential to leverage information from multiple modalities before making the final prediction~\cite{antol2015vqa}. 
Hence, it is crucial while challenging to develop multi-modal PWS methods to improve the data efficiency on these tasks.

\paragraph{Extend the Scope of Usable LFs.}
Although researchers have made attempts to extend the scope of usable LFs~\cite{zhang2021creating,yu2021learning}, there are other sources that could potentially be used as LFs, \eg, physical rules, for more complex tasks. The ultimate goal of PWS is to leverage as more existing sources as possible to minimize human efforts in the curation of training data.

\paragraph{Ethical and Trustworthy AI.}
One of the most pressing concerns in the AI community right now is ensuring that AI techniques and models are applied ethically.
Within this area of focus, one of the most important and challenging topics is ensuring that the training data which informs models is ethically labeled and managed, transparent, auditable, and bias-free.
PWS approaches offer a step-change opportunity in this regard, since they result in training labels generated by code which can be inspected, audited, governed, and edited to reduce bias.
However, by the same token, PWS methods can also lead to more direct bias in training data sets if used and modeled improperly~\cite{geva2019modeling,lucy2021gender}.
Overall, further systematic study in this area is highly critical, and has great opportunity for improving the state of data in AI from an ethics and governance perspective.

\section{Conclusion}

Manual annotations are always of great importance to training machine learning models, but usually expensive and time-consuming. Programmatic weak supervision (PWS) offers a promising direction to achieve large-scale annotations with minimal human efforts. In this article,  we review the PWS area by introducing existing approaches for each component inside a PWS workflow. We also describe how PWS could interact with methods from related fields for better performance on downstream applications. Then, we list existing datasets and recent applications of PWS in the literature. Finally, we discuss current challenges and future directions in the PWS area, hoping to inspire future research advances in PWS.

\bibliographystyle{named}
\bibliography{ijcai22}

\end{document}